\documentclass[10pt,journal]{IEEEtran}
\usepackage{tikz}
\newcommand*\circled[1]{\tikz[baseline=(char.base)]{
       \node[shape=circle,fill,inner sep=1pt] (char) {\textcolor{white}{\small#1}};}}

\ifCLASSOPTIONcompsoc
  \usepackage[nocompress]{cite}
\else
  \usepackage{cite}
\fi

\usepackage{epsfig}
\usepackage{verbatim}
\usepackage{balance}
\usepackage{float}
\usepackage{algorithm}
\usepackage{amssymb}
\usepackage[english]{babel}
\usepackage{array}
\usepackage{comment}
\usepackage{multirow}
\usepackage{multicol}
\usepackage{comment}
\usepackage{enumerate}
\usepackage{color}
\usepackage{listings}
\usepackage{graphicx}
\usepackage{epstopdf}
\usepackage{subfigure}
\usepackage{enumitem}
\usepackage{url}
\usepackage{textcomp}
\usepackage{amsmath}
\usepackage{amsmath,amssymb,amsfonts}
\usepackage{graphicx}
\usepackage{textcomp}
\usepackage{xcolor}
\usepackage{cleveref}
\usepackage{lipsum}
\usepackage[utf8]{inputenc}
\usepackage[noend]{algpseudocode}
\usepackage{tabularx}

\newcommand{\safepath}{\emph{SAfEPaTh}}

\newcommand*{\rom}[1]{\uppercase\expandafter{\romannumeral #1\relax}}

\newenvironment{conditions*}
{\par\vspace{\abovedisplayskip}\noindent
	\tabularx{\columnwidth}{>{$}l<{$} @{${}={}$} >{\raggedright\arraybackslash}X}}
{\endtabularx\par\vspace{\belowdisplayskip}}

\begin{document}

\title{SAfEPaTh: A System-Level Approach for Efficient Power and Thermal Estimation of Convolutional Neural Network Accelerator}

\author{
    \IEEEauthorblockN{Yukai Chen\IEEEauthorrefmark{1}, Simei Yang\IEEEauthorrefmark{1}, Debjyoti Bhattacharjee\IEEEauthorrefmark{1}, Francky Catthoor\IEEEauthorrefmark{1}, Arindam Mallik\IEEEauthorrefmark{1}} \\
    \IEEEauthorblockA{\IEEEauthorrefmark{1}IMEC, Leuven, Belgium} \\
    Email: yukai.chen@imec.be, debjyoti.bhattacharjee@imec.be, arindam.mallik@imec.be}

\IEEEtitleabstractindextext{%
\begin{abstract}
The design of energy-efficient, high-performance, and reliable Convolutional Neural Network (CNN) accelerators involves significant challenges due to complex power and thermal management issues. This paper introduces SAfEPaTh, a novel system-level approach for accurately estimating power and temperature in tile-based CNN accelerators. By addressing both steady-state and transient-state scenarios, SAfEPaTh effectively captures the dynamic effects of pipeline bubbles in interlayer pipelines, utilizing real CNN workloads for comprehensive evaluation. Unlike traditional methods, it eliminates the need for circuit-level simulations or on-chip measurements.
Our methodology leverages TANIA, a cutting-edge hybrid digital-analog tile-based accelerator featuring analog-in-memory computing cores alongside digital cores. Through rigorous simulation results using the ResNet18 model, we demonstrate SAfEPaTh's capability to accurately estimate power and temperature within 500 seconds, encompassing CNN model accelerator mapping exploration and detailed power and thermal estimations. This efficiency and accuracy make SAfEPaTh an invaluable tool for designers, enabling them to optimize performance while adhering to stringent power and thermal constraints. Furthermore, SAfEPaTh's adaptability extends its utility across various CNN models and accelerator architectures, underscoring its broad applicability in the field. This study contributes significantly to the advancement of energy-efficient and reliable CNN accelerator designs, addressing critical challenges in dynamic power and thermal management.

\end{abstract}


}

\maketitle

\IEEEdisplaynontitleabstractindextext
\IEEEpeerreviewmaketitle

\ifCLASSOPTIONcompsoc
\IEEEraisesectionheading{\section{Introduction}\label{sec:introduction}}
\else
\section{Introduction}
\label{sec:introduction}
\fi
Convolutional Neural Network (CNN) accelerators have become essential for many modern applications, such as autonomous driving, image and speech recognition, and natural language processing, all of which demand low power consumption, high performance, and high reliability~\cite{capra2020updated}. However, as semiconductor scaling approaches its physical limits, managing dynamic power and thermal issues in accelerators executing CNN models poses increasing challenges. Accurate power and thermal estimation methods are crucial to mitigate the adverse effects of high temperatures and power consumption while ensuring the device operates within its thermal constraints.

To assess the power and thermal characteristics of CNN accelerators, hardware-based built-in chip sensors are commonly employed~\cite{chen2022thermal}. Despite their utility, these sensors exhibit limited resolution and are unsuitable for extensive design space exploration, leading to substantial challenges in power and thermal estimation. Consequently, simulations have emerged as a valuable tool for studying and understanding the thermal and power behavior of CNN accelerators. Simulations facilitate comprehensive evaluations of CNN accelerator designs across a broader exploration range, considering performance, energy efficiency, and the power and thermal attributes of the system. As technology advances, refined thermal management becomes crucial to ensure the device operates within its thermal constraints and to reduce leakage power. Thus, power and thermal estimation are foundational for designing and implementing effective management policies.

Recent research has highlighted thermal issues as a critical concern for CNN accelerators. Existing works primarily focus on the thermal problems of specific components, such as ReRAM-based Analog-in-Memory Computing (AiMC) arrays~\cite{chen2022wrap} or systolic arrays~\cite{sze2020evaluate}. In contrast, this work examines the entire system of tile-based CNN accelerators. These accelerators typically consist of a 2D array of tiles, where each tile houses Processing Elements (PEs) functioning as data-parallel vectorized or array-based processor templates. These tiles share an L1 memory serving as a scratchpad rather than a hardware-controlled cache. Tile-based CNN accelerators support both intra-layer parallelism and inter-layer pipelining. However, due to inter-layer data dependencies~\cite{song2017pipelayer}, these accelerators are prone to pipeline bubbles, potentially leading to execution stalls. In such systems, steady-state power and thermal metrics provide only a coarse estimation of the overall system performance, limited to capturing the pipeline execution of a CNN workload. Therefore, it is essential to consider the power consumption and thermal distribution of tile-based CNN accelerators at a finer granularity, enabling thermal-aware Design Space Exploration (DSE) that includes detailed transient responses.

In this paper, we propose \safepath, a system-level methodology for power and thermal estimation of tile-based CNN accelerators, employing detailed power and thermal models in both steady-state and transient-state scenarios. The main contributions of our work are as follows:
\begin{itemize}
    \item A novel transient power trace extraction method is applicable at various granularities, including pixel-, layer-, and inference-granularity. The generated power trace reflects the execution of a CNN inference workload mapped onto a tile-based accelerator.
    \item By adapting the power trace extracted from realistic CNN workloads, we conduct a thermal upper bound analysis of the CNN accelerator.
\end{itemize}

Our methodology obviates the need for circuit-level simulations and on-chip measurements, facilitating fast and efficient DSE with transient responses for CNN accelerators. In Section~\ref{sec:back_related}, we provide background information on power and thermal analysis of CNN accelerators and thermal estimation in EDA fields. Section~\ref{sec:methods} details our proposed methodology, while Section~\ref{sec:results} presents the power and thermal simulation results and related analysis. Finally, in Section~\ref{sec:conclusion}, we summarize our work and discuss future research directions.

\section{Background and Related Works} \label{sec:back_related}
\subsection{Tile-Based CNN Accelerators}
\begin{figure}[!thbp]
\begin{center}
\includegraphics[width=0.90\linewidth]{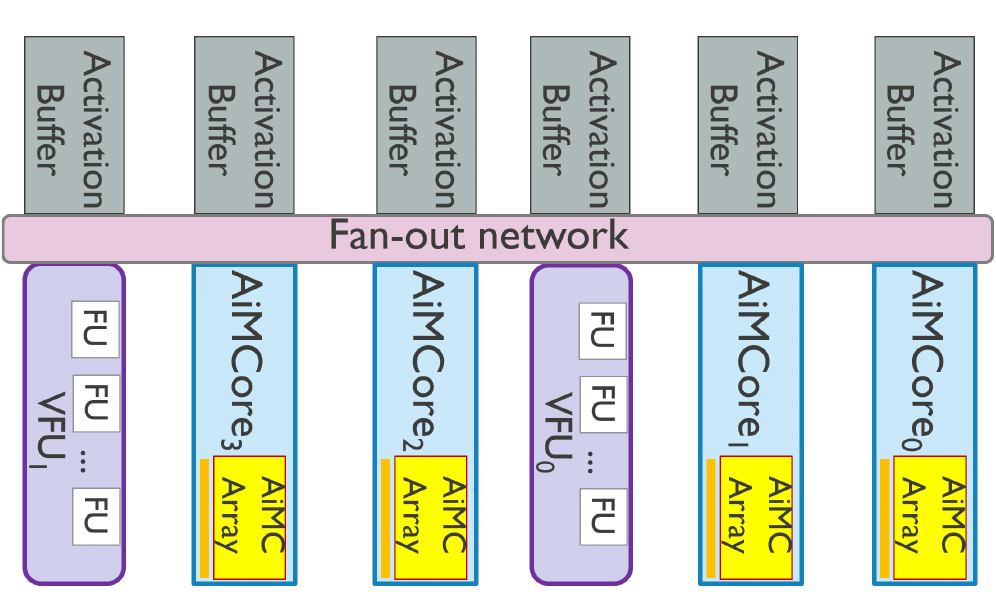}
\caption{TANIA tile-level architecture, consisting of PEs (AiMCore with AiMC arrays~\cite{cosemans2019towards}, digital Vector Functional Units (VFUs)), distributed activation buffers (ActBufs), and fanout network for intra-tile communication~\cite{yang2022aero}.}
\label{fig:tania}
\end{center}
\end{figure}

Tile-based accelerators have gained considerable attention due to their scalability, high performance, and energy efficiency. These accelerators typically consist of multiple tiles interconnected through a Network-on-Chip (NoC)~\cite{shafiee2016isaac,song2017pipelayer,yang2022aero}. Each tile contains one or more spatial arrays, such as PEs and/or Analog-in-Memory Computing (AiMC) cores (referred to as AiMCore)~\cite{cosemans2019towards}, along with on-chip memories. Tile-based accelerators support both intra-layer parallelism and inter-layer pipelining of CNN models, enabling multiple layers of CNNs to execute across tiles in a pipelined fashion, thus reducing data transfers between on-chip memory and off-chip DRAM. However, pipeline bubbles can occur due to inter-layer data dependencies, resulting in computational component stalls. Figure~\ref{fig:tania} illustrates one tile of the TANIA architecture~\cite{yang2022aero}, which serves as a demonstrator in this work.

Most DSE frameworks~\cite{shafiee2016isaac,yang2022aero} have focused on optimizing the throughput, latency, resource utilization, and energy efficiency of CNN accelerators. Previous studies, such as~\cite{yang2022aero}, explored the mapping of a tile-based accelerator with Performance, Power, and Area (PPA) profiling evaluation, reporting energy efficiency in joules per inference or Tera-Operations per Second per Watt (TOPS/W). However, most existing accelerator DSE frameworks do not consider power consumption, measured in watts (W), as an exploration metric and primarily rely on energy estimation, measured in Joules (J) or millijoules (mJ), as a proxy for power consumption. This is because power consumption across inference latency does not provide more information than joules per inference when designing an energy-efficient system. Nonetheless, high power consumption can result in elevated temperatures, potentially causing device failure, degraded performance, shortened lifetime, compromised reliability, and security issues.

Fine-grained power estimation is challenging and can increase simulation overhead, but it is a critical metric for designing high-performance, thermally-sensitive CNN accelerators. Consequently, a system-level methodology is necessary to perform fast, fine-grained power and thermal estimation for tile-based accelerators, including transient analysis to capture both spatial and temporal variations. Such a methodology enables designers to comprehensively understand the power and thermal behaviors under various workloads and conditions, facilitating the development of robust and efficient CNN accelerators that meet stringent performance and reliability requirements.

\subsection{Power and Thermal Issues of CNN Accelerators}
In recent years, thermal issues of CNN accelerators have emerged as a significant concern. Prior work~\cite{chen2022wrap} focuses on thermal-aware optimization for ReRAM-based DNN accelerators, where ReRAM serves as an AiMC component for multiply-accumulate operations. This work aims to reduce the average temperature and temperature variation of ReRAM arrays to maintain training accuracy. The study achieves this by employing weight mapping optimization methods such as column reordering, weight splitting, and weight compensation. Additionally, analytical models are used to obtain temperature variance based on the weight mapping of the ReRAM array. 

In contrast, the work~\cite{zervakis2022thermal} examines approximate multipliers in a systolic array to mitigate on-chip thermal issues. Another notable study~\cite{shukla2021temperature} presents a thermal-aware optimization framework for Mono3D CNN accelerators (systolic array-based). Their architecture optimization flow begins with a performance evaluation simulator, where the output (e.g., memory requests), combined with an SRAM simulator and Mono3D power models, is used to obtain power traces. The \emph{Hotspot} simulator is then applied to determine the steady-state temperatures of the accelerator.

Our work is closely related to~\cite{shukla2021temperature}, as we also aim to generate power traces of the system to feed into the \emph{Hotspot} thermal estimation tool. However, our focus is on tile-based accelerators. More importantly, we consider both power and thermal estimates in the steady state and the transient state based on realistic workloads to evaluate the system's performance over time. This approach allows us to capture the dynamic thermal behavior of the accelerators, providing a more comprehensive understanding of their performance and reliability under varying operational conditions.

\subsection{Thermal Estimation Methods}
The advancement of technology and the miniaturization of device dimensions have made thermal estimation a crucial aspect of the system design process to achieve optimal performance and prevent thermal failures. Recent thermal estimation works~\cite{zhang2021full} employ machine learning and deep learning techniques to conduct thermal estimation during system runtime. However, these methods are on an ad-hoc basis, and the trained models are typically not transferable. Additionally, creating the training dataset, composed of infrared thermal images recorded by infrared cameras, demands significant effort and renders these works inapplicable to the system's design phase.

The advancement of technology and the miniaturization of device dimensions have made thermal estimation a crucial aspect of the system design process, essential for achieving optimal performance and preventing thermal failures. Recent thermal estimation works~\cite{zhang2021full} employ machine learning and deep learning techniques to conduct thermal estimation during system runtime. While innovative, these methods are often ad-hoc, and the trained models are typically not transferable to different systems or conditions. Additionally, creating the training datasets, which are composed of infrared thermal images recorded by infrared cameras, demands significant effort and renders these methods impractical for use during the system design phase.

The \emph{Hotspot} framework~\cite{huang2006hotspot} is widely recognized in the EDA research community for architecture-level thermal exploration and management of circuits and systems. Numerous thermal simulators are built upon the thermal models used in \emph{Hotspot}, though some employ different kernel functions to speed up the resolution of differential equations and enhance performance~\cite{chen2018systemc,jiang2022chip}. Certain simulators are specifically designed for 3D chips and modify the thermal model structure to accommodate this application~\cite{sridhar20133d,siddhu2022comet}. Other approaches have implemented new methodologies to capture the power-temperature feedback loop during concurrent simulation~\cite{siddhu2022comet,chen2018systemc}. 

However, it is important to note that many of these thermal simulators use synthetic or fixed random constant power values as input, which fails to capture the thermal behavior under real workloads accurately. This limitation can lead to suboptimal thermal management strategies that do not reflect actual operating conditions. Therefore, there is a critical need for methodologies that can accurately estimate power and temperature based on realistic, dynamic workloads, particularly during the design phase. Our proposed methodology addresses this gap by providing a system-level approach for fine-grained power and thermal estimation, including both steady-state and transient-state scenarios, using real CNN workloads. This enables more accurate and effective thermal management strategies, ensuring the reliability and performance of CNN accelerators in practical applications.

\section{Proposed Methodology} \label{sec:methods}
\begin{figure}[!thbp]
\begin{center}
\includegraphics[width=0.99\linewidth]{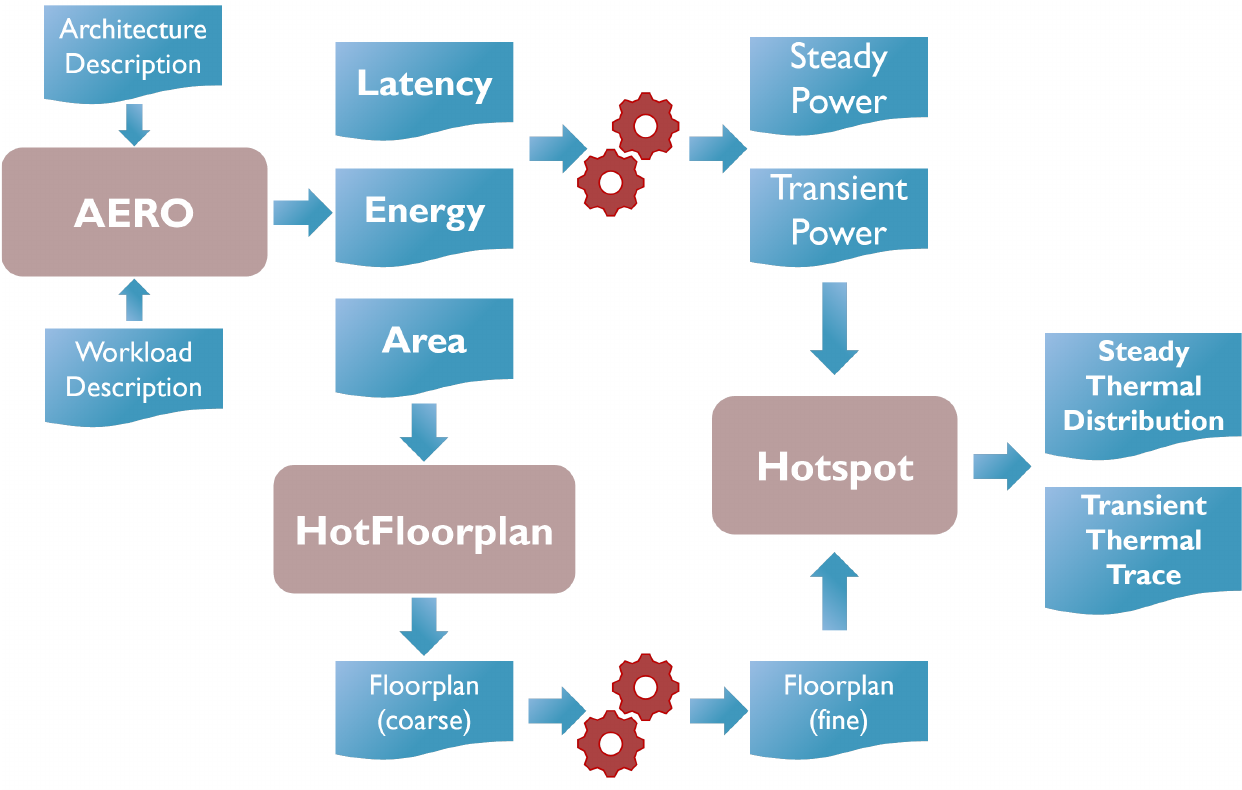}
\caption{Proposed system-level power and thermal estimation methodology.}
\label{fig:method}
\end{center}
\end{figure}

Figure~\ref{fig:method} provides a detailed depiction of our proposed system-level power and thermal estimation methodology, which comprises three primary tools: \emph{AERO}, \emph{HotFloorplan}, and \emph{Hotspot}. The gear icons shown in Figure~\ref{fig:method} signify the crucial data processing steps involved in our proposed approach. The detailed process to acquire both steady-state and transient power and temperature values is described below.

\subsubsection{\bf Latency, Energy, and Area Extraction}
Our methodology begins with \emph{AERO}~\cite{yang2022aero}, a mapping and profiling framework that extracts latency, energy, and area estimations. The framework takes a CNN inference workload and the TANIA architecture, which we use in our experimental case study, as inputs and explores intra-layer parallelism and inter-layer pipeline mapping. With the mapping solution, \emph{AERO} generates a preliminary set of instructions for each PE, assuming ideal data placement, where the data required for each computation is continuously available.

\emph{AERO} models each pixel computation's start time and finish time at pixel granularity, where the average number of cycles executed by instructions to generate one-pixel output characterizes pixel computation time, ignoring corner cases such as padding. Figure~\ref{fig:powerTrace} (a) shows an example of latency characterization for three CNN layers (see \circled{\small{\bf{1}}}). Since there is inter-layer data dependency, computation for a pixel in a layer only starts after all the required data is available from the input layers. This can result in pipeline bubbles, as highlighted in the figure. \emph{AERO} also supports on-chip communication latency estimation, taking into account intra-tile communication cycles in instructions.
 
\emph{AERO} uses {\em Accelergy}\cite{wu2019accelergy}, a component-level framework, to estimate the area and energy of a given workload mapped to the hardware architecture (see \circled{\small{\bf{2}}} in Figure\ref{fig:powerTrace}). {\em Accelergy} models the hardware architecture as component classes, where each class has a component-level estimator plugin that specifies the area and energy consumption of different actions. For example, the component classes in a TANIA-tile (Figure~\ref{fig:tania}) include ActBuf, AiMCore, and a compound component containing the SIMD (Single Instruction Multiple Data) lanes in a VFU~\cite{yang2022aero}. By combining the estimator plugins of each component with runtime action counts generated from instructions of a specific CNN workload (\circled{\small{\bf{3}}} in Figure\ref{fig:powerTrace}), area and energy consumption can be estimated.

\subsubsection{\bf Transient Power Trace Calculation}
\begin{figure}[!htbp]
\begin{center}
\includegraphics[width=1.66\linewidth]{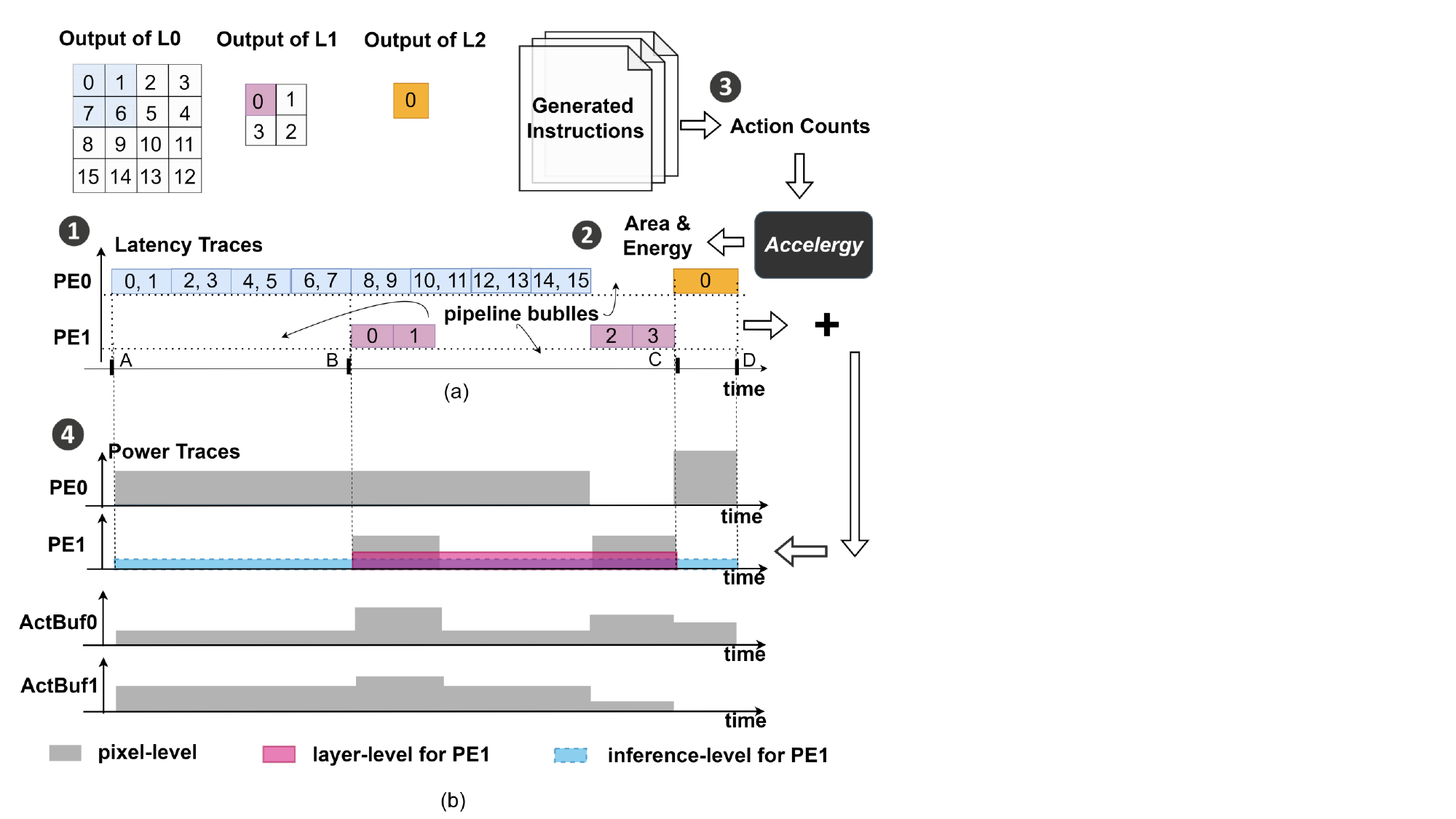}
\caption{(a) Latency, area, and energy estimation of 3 CNN-layers mapped onto 2 PEs by using AERO; (b) Pixel-level power traces for different PEs and ActBufs, layer-level and inference-level power traces for PE1 (Assumes data is read from ActBuf0 and is written to ActBuf1 for computing L0).}
\label{fig:powerTrace}
\end{center}
\end{figure}

Our proposed approach takes latency traces and instructions obtained from \emph{AERO} as inputs to generate power traces at different granularities (e.g., pixel-level, layer-level, inference-level). Figure~\ref{fig:powerTrace} (b) provides an example of power trace generation at the pixel level. First, we process the instructions for each output pixel to obtain the corresponding action counts for hardware components (e.g., PE, ActBuf, AiMCore). We then pass the action counts to {\em Accelergy} to obtain the energy consumption and calculate the pixel-level power based on the pixel latency (${Power}=\frac{Energy}{Latency}$). Finally, we combine the power values on the y-axis of the latency traces to obtain power traces for different hardware components (see \circled{\small{\bf{4}}}).

Similarly, we can obtain layer-level and inference-level power traces by computing average power within layer latency and inference latency, respectively (e.g., PE1 in Figure~\ref{fig:powerTrace} (b)). By using this approach, we can also obtain instruction-level/cycle-level power traces with finer-grained data processing. In this work, to study the impact of pipeline bubbles in tile-based architectures, we use pixel-level power traces to estimate transient power and inference-level power traces to estimate steady-state temperature.

\subsubsection{\bf System-level Floorplan Generation}
Estimating the temperature of a system requires a reasonable floorplan, which is challenging to obtain at the front end of the system design. In our proposed methodology, we use \emph{AERO} to extract the area information of each component in the system and generate a single tile floorplan using \emph{Hotflooplan}, the built-in floorplan generation tool provided by \emph{Hotspot}. From this single tile, we generate multiple tiles and a cluster floorplan. However, obtaining a reasonable floorplan for each tile with numerous identical components takes time and effort. To address this issue, we propose a method that reduces the number of components by combining identical components into one lumped component. Each lumped component's maximum and minimum height and width ratio is set for \emph{Hotflooplan}, and the floorplan evaluation cost function is updated to control the adjacency information among lumped components. This approach enables us to obtain a coarse-grained floorplan, which we then refine into a fine-grained floorplan by cutting the lumped components into identical components based on the same type number. A detailed floorplan generation process is indicated in Section~\ref{sec:results}.

\subsubsection{\bf Steady-State and Transient Thermal Estimation}
To estimate the steady-state and transient temperature, we adopt the \emph{Hotspot} tool in our methodology. In the previous step, a fine-grained floorplan is generated, which includes the name, size, and coordinates of each component. This floorplan serves as one of the input files for \emph{Hotspot}. The other input file is the power traces file, which contains the power consumption of each component in the accelerator of the CNN workload at each simulation step. Using these input files, \emph{Hotspot} performs thermal simulations and generates steady-state thermal distribution and transient temperature traces under specific physical and environmental configurations based on real CNN workloads.

\section{Experiment Results} \label{sec:results}
\subsection{Experimental Simulation Setup}
\subsubsection{System Architecture Description}
In this work, we use TANIA as our underlying tile-based CNN accelerator and focus on the power and thermal estimation of a single tile to demonstrate the effectiveness of the proposed methodology. Future research will explore power and thermal aspects in multi-tile and multi-cluster scenarios. Figure~\ref{fig:tania} illustrates the TANIA-tile architecture, consisting of 4 analog AiMCores and 2 digital VFUs. According to~\cite{yang2022aero}, we set the size of the AiMCore to be 1,152 rows $\times$ 512 columns. Each ActBuf and Instruction Memory (IMem) within each AiMCore and VFU are 1,536KB and 128KB, respectively. AiMCore operates at 0.1GHz, while VFU operates at 1GHz.

\subsubsection{AERO framework configuration}
The ability of \emph{AERO} to conduct design space exploration under resource constraints is a significant advantage compared to other existing design space exploration frameworks. This feature enables the mapping of all workloads onto a single tile, which allows us to focus on the power and thermal analysis within the single tile using our proposed approach, disregarding inter-communications among different tiles and clusters. 
In the following simulations, we configure the TANIA only with one available tile when we use \emph{AERO} to conduct the design space exploration to obtain the latency, energy, and area results as described in Section~\ref{sec:methods}. 

\subsubsection{Hotspot Thermal Configuration}
Table~\ref{table:hotspot} presents the primary physical parameters of the packaging layers utilized in \emph{Hotspot}. The air and heat sink convection resistance are set to $0.17 K/W$, and we use an ambient temperature of 40$^{\circ}$C (313.15K) in all subsequent thermal simulations.

\begin{table}[!htbp]
\centering
\caption{Physical parameters of packaging}\label{table:hotspot}
\begin{tabular}{|c|r|r|r|}
\hline
Name          & \multicolumn{1}{c|}{Size (mm)} & \multicolumn{1}{l|}{Thickness (mm)} & \multicolumn{1}{l|}{$K_{th}$ (W/mK)} \\ 
\hline\hline
Silicon Chip                    & $2.261 \times 2.242$                          & 0.5                                       & 140                               \\ \hline
TIM                    & $2.261 \times 2.242$                          & 0.1                                     & 7                                 \\ \hline
Heat Spreader          & $3.375 \times 3.375$                        & 0.2                                     & 400                               \\ \hline
Heat Sink              & $4.500 \times 4.500$                            & 1.0                                       & 400                               \\ \hline
\end{tabular}
\end{table}

\subsection{One-Tile Floorplan of TANIA}
\begin{figure}[!thbp]
\begin{center}
\includegraphics[width=0.98\linewidth]{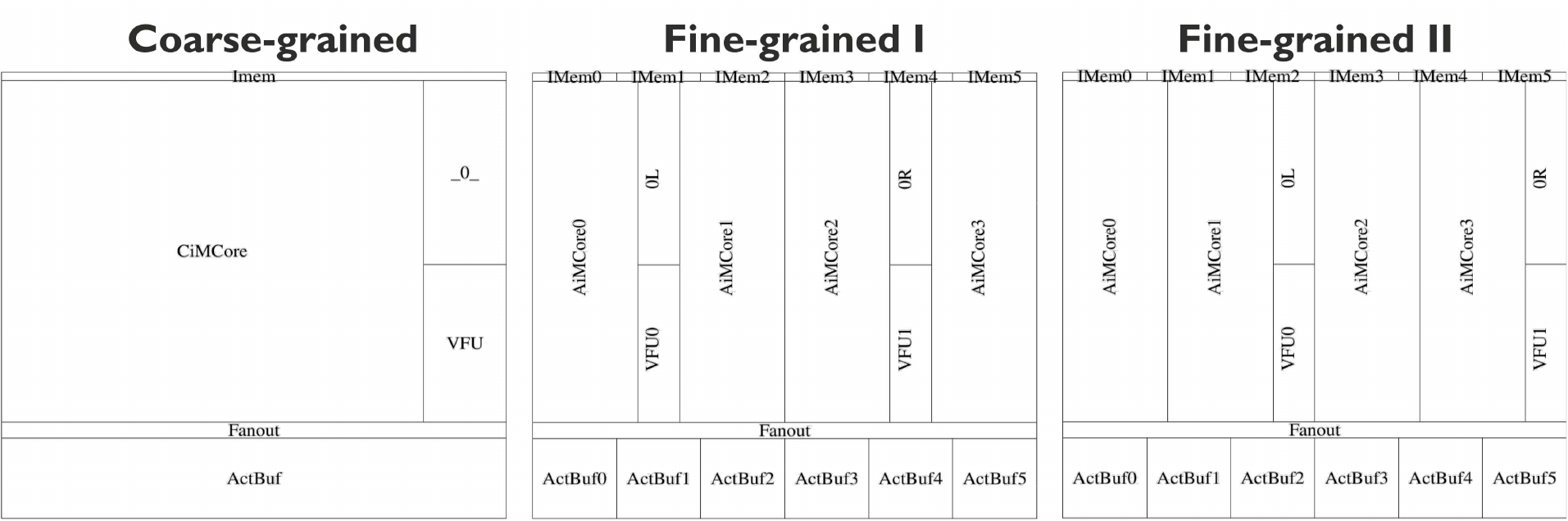}
\caption{Floorplan of one tile in TANIA: Coarse-grained and fine-grained.}
\label{fig:taniafloorplan}
\end{center}
\end{figure}

Figure~\ref{fig:taniafloorplan} shows the floorplan of a single tile in TANIA, including both coarse-grained and two fine-grained versions generated using the methodology described in Section~\ref{sec:methods}. 
The entire tile measures ${\mbox 2.26mm \times 2.24mm}$, which is also used as the silicon die size parameter in \emph{Hotspot}. The two fine-grained floorplans differ only in the VFUs' coordinates.

\subsection{Power and Thermal Simulation Results}

\subsubsection{Power Traces}
\begin{figure}[!thbp]
\begin{center}
\includegraphics[width=0.99\linewidth]{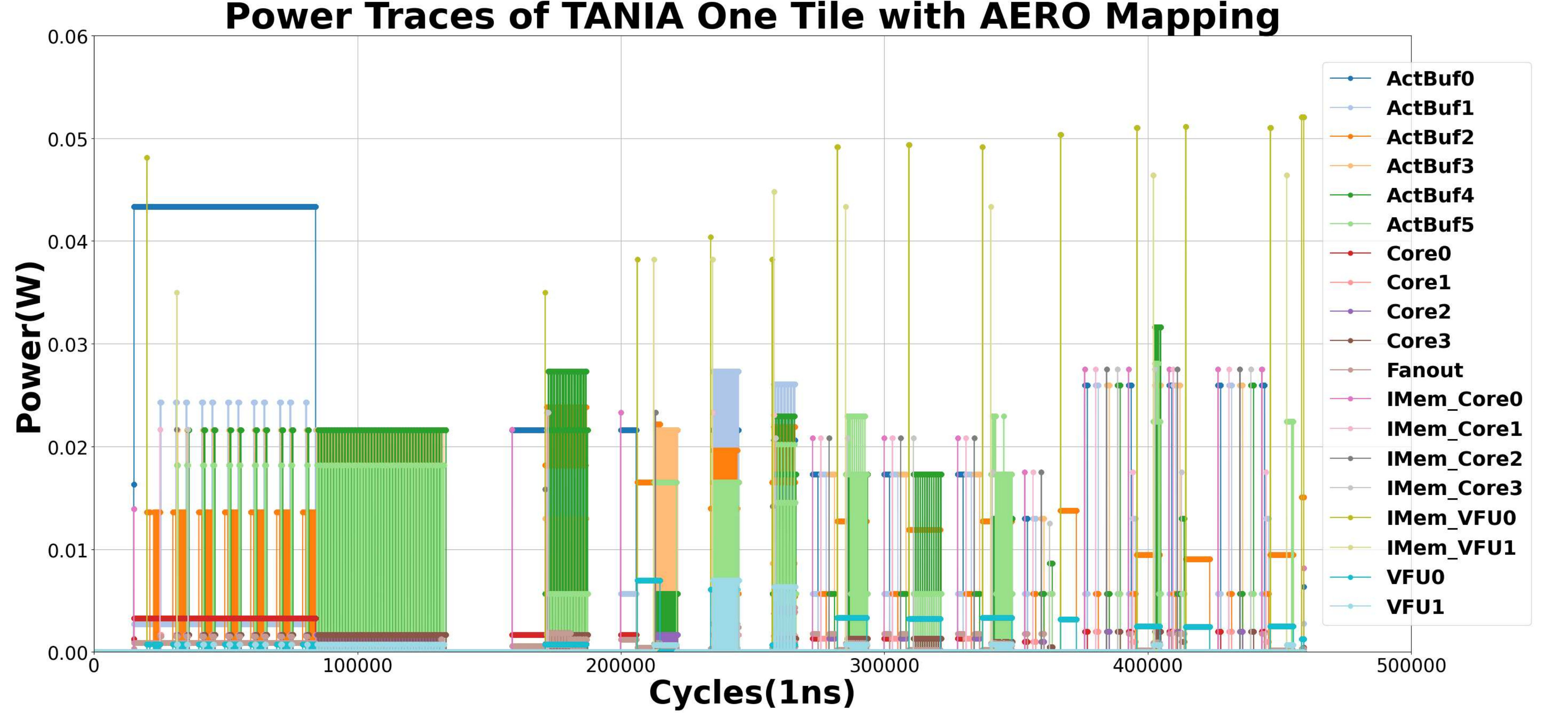}
\caption{Power traces of TANIA runs Resnet18 under AERO mapping.}
\label{fig:poweraero}
\end{center}
\end{figure}

Using the proposed methodology, we obtained power traces for all units within one tile of TANIA, with the Resnet18 model's inference as the underlying workload. Figure~\ref{fig:poweraero} presents the power traces associated with the {\em AERO's} mapping results, as described in Section~\ref{sec:methods}. The total energy consumption for one Resnet18 model's inference is $12.308uJ$, and the total latency is 459,239 cycles. To minimize energy consumption, each PE (VFU/AiMCore) in TANIA fetches the required instructions once from the IMem into loop buffers. The PEs then execute instructions from the loop buffers, which result in negligible energy consumption \cite{jayapala2005clustered}. As a result, the power consumption of different IMems exhibited many peaks, as shown in Figure~\ref{fig:poweraero}.

\subsubsection{Thermal Steady-State Distribution and Transient Traces}
\begin{figure}[!thbp]
\begin{center}
\includegraphics[width=0.99\linewidth]{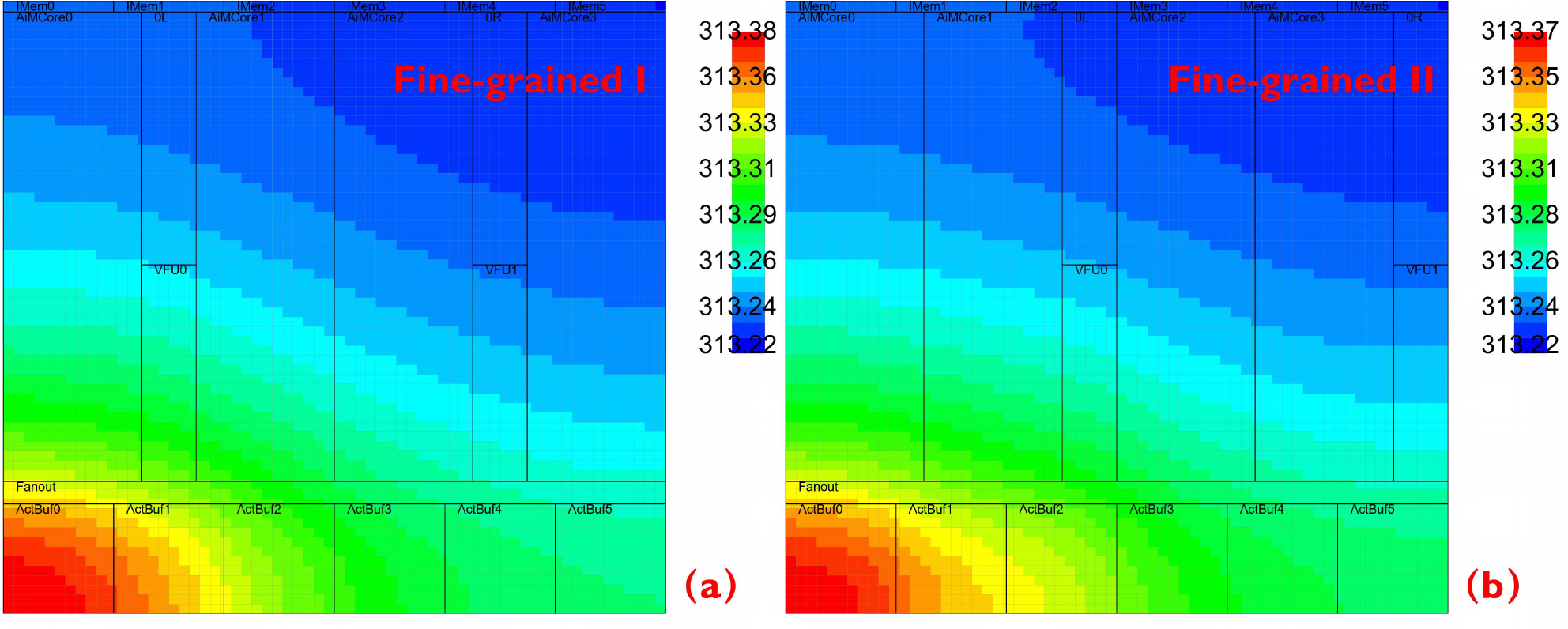}
\caption{Steady-state thermal distribution of TANIA runs Resnet18 model with AERO mapping.}
\label{fig:stempaero}
\end{center}
\end{figure}

\begin{figure}[!thbp]
\begin{center}
\includegraphics[width=0.99\linewidth]{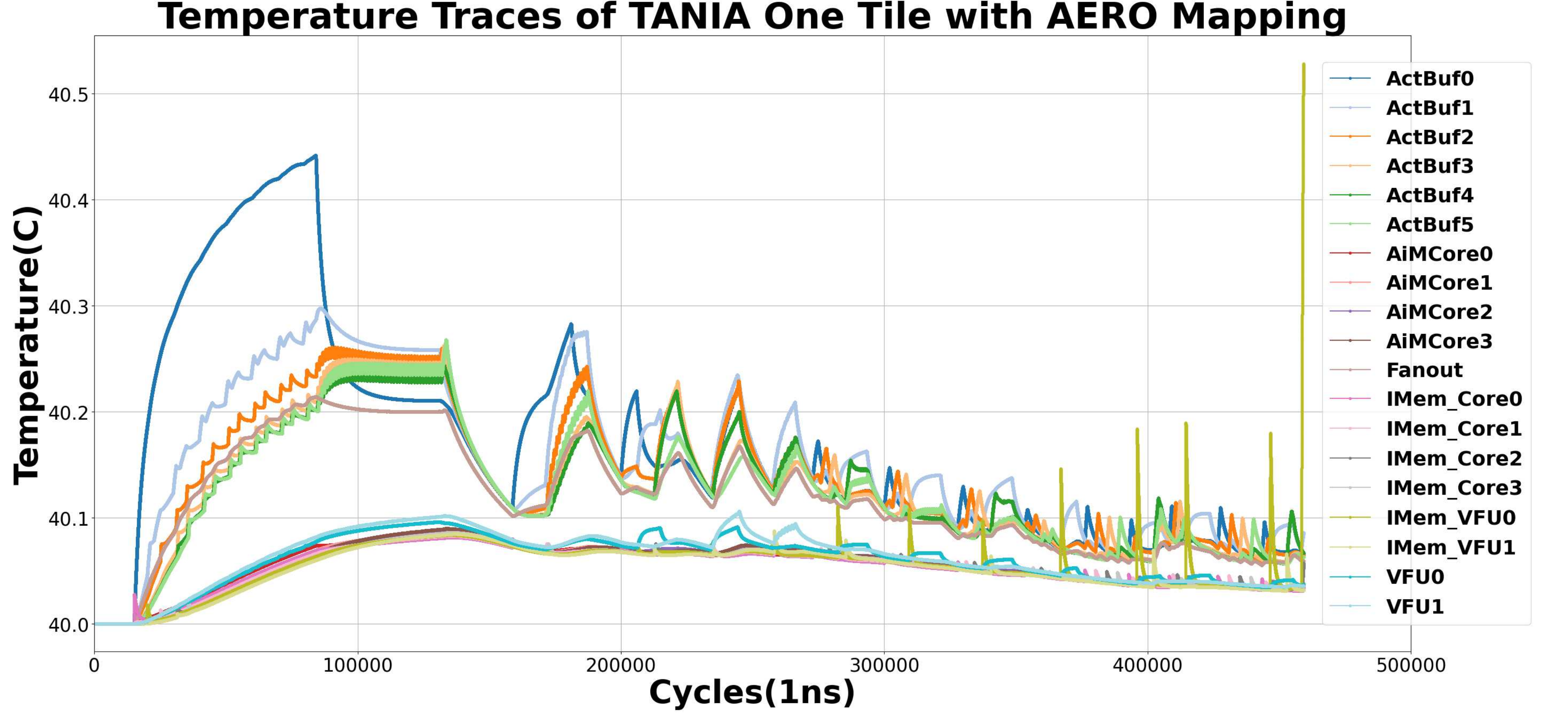}
\caption{Temperature traces of TANIA runs Resnet18 with AERO mapping.}
\label{fig:ttempaero}
\end{center}
\end{figure}

Figure~\ref{fig:stempaero} shows the steady-state thermal distribution of two fine-grained one-tile floorplans under the power traces obtained from the {\em AERO}, as shown in Figure~\ref{fig:poweraero}. The results demonstrate that the floorplans exhibit negligible differences in their thermal distribution, with the first ActBuf associated with the first AiMCore being the hottest component and the northeast corner being the coolest spot. The first ActBuf's elevated temperature is attributed to the {\em AERO's} assumption that the input image for the CNN model is ready, allowing the first layer's computation to be continuous without interruption, and its computation consumes the most energy due to its large pixel count compared with the following layers. Additionally, the ActBuf continuously passes the first layer's computed intermediate results to the second layer without stopping, leading to an uninterrupted temperature increase, as indicated by the continuous blue line in Figures~\ref{fig:poweraero} and~\ref{fig:ttempaero}. The maximum temperature difference between the two floorplans is only 0.01$^{\circ}$C, and it is important to note that when scaling this floorplan to a smaller technology node, the absolute temperatures will increase, although never as high as for digital cores with equivalent performance levels. This result indicates that the VFUs' location has a limited impact on the thermal distribution. Therefore, all subsequent simulations are based on the second fine-grained floorplan.

Figure~\ref{fig:ttempaero} displays the transient temperature traces of each component in the TANIA tile. These traces were computed by feeding the power traces of TANIA running Resnet18 into Hotspot, as shown in Figure~\ref{fig:poweraero}. The temperature increase resulting from one Resnet18 inference is limited, less than 0.5$^{\circ}$C. This limitation is due to the TANIA tile's relatively large area, which is attributed to the existence of AiMCores, making it approximately ten times larger than traditional pure digital accelerators. This results in a low power density. Furthermore, AiMCores consume less power per operation than digital cores, which reduces the power density even more. Additionally, non-ideal mapping policies and dependencies among different layers in the CNNs cause ``bubbles" in the power traces, as described in Section~\ref{sec:methods}. These bubbles correspond to time intervals when most components in the TANIA tile are inactive, such as the interval between 150,000 cycles and 180,000 cycles in Figure~\ref{fig:poweraero} and~\ref{fig:ttempaero}. During these inactive time intervals, inactive and neighboring components cool down, preventing significant temperature increases. Eliminating these bubbles from the power traces results in higher performance mapping power traces, which can even achieve the ideal mapping results. Analyzing the temperature using these improved power traces without ``bubble" periods can obtain an upper bound on the temperature rise for the entire inference process. A case study of this upper-bound temperature analysis is presented in Section~\ref{sec:casestudyupper}.

\subsection{Thermal Analysis Case Studies}
The proposed methodology demonstrates remarkable efficacy, as it enables a fully automated process encompassing hardware mapping, power calculation, and temperature analysis, which can be completed within 500 seconds with the ResNet18 model serving as the representative workload. This rapid process allows for swift DSE while considering the impacts of temperature and power.

\subsubsection{Thermal-Aware Mapping Analysis}
\begin{figure}[!thbp]
\begin{center}
\includegraphics[width=0.99\linewidth]{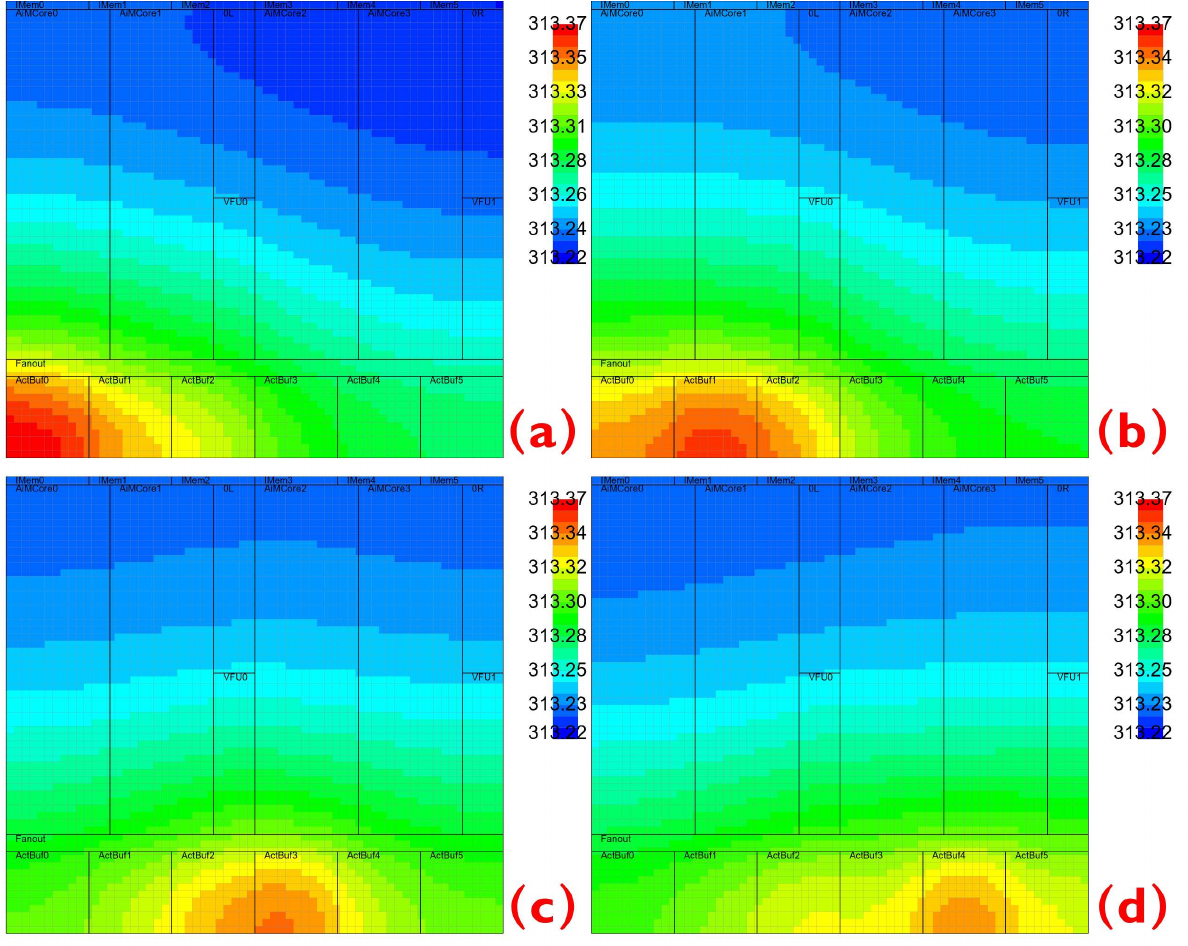}
\caption{Steady-state thermal distribution of TANIA's fine-grained floorplan \rom{2} run Resnet18 under different workload mappings.}
\label{fig:thermaldisvar}
\end{center}
\end{figure}

Figure~\ref{fig:stempaero} reveals that the leftmost ActBuf is the hottest component in the TANIA architecture due to {\em AERO} allocating the first convolutional layer to its associated PE. To investigate the impact of workload mapping on the architecture's thermal behavior, we conducted a thermal-aware mapping exploration. Specifically, we switched the hottest PE and its corresponding ActBuf and IMem with the coldest PE and its corresponding ActBuf and IMem, while keeping the other component locations unchanged. Figure~\ref{fig:thermaldisvar} illustrates the four cases of exchanging the workload (layer of CNN) of the first AiMCore with the second, third, and fourth AiMCore, respectively. This analysis provides insights into the thermal distribution of the TANIA architecture and the impact of workload mapping on its thermal behavior.

To compare the differences in thermal distribution resulting from the thermal-aware mapping exploration, we plotted the thermal distributions for all four cases using the same thermal scale. As shown in subplot (d) of Figure~\ref{fig:thermaldisvar}, exchanging the workload between the first and last AiMCore resulted in the smallest maximum temperature for the tile. This mapping separated the most active components of the entire tile at a relative distance, leading to the smallest temperature increase of 0.19$^{\circ}$C, a 13.63\% reduction from the maximum temperature increase of 0.22$^{\circ}$C (subplot (a) of Figure~\ref{fig:thermaldisvar}). This gain in thermal performance is attributed to the mapping that distributes the workload more evenly in space. It is important to note that while the AiMCores in the TANIA tile contribute to the limited absolute temperature increase, the relative trends still show the potential impact of mapping.

\subsubsection{Transient Temperature Upper Bound Analysis}\label{sec:casestudyupper}
The power traces obtained from executing CNNs on TANIA contain ``bubble" periods, which provide opportunities for the tile to cool down. This makes it challenging to determine the upper bound of the transient temperature during CNN inference. To obtain an upper bound on the instantaneous temperature, we revised these power traces to mimic those acquired from high-performance-focused mapping tools and even ideal scenarios.

\begin{figure}[!thbp]
\begin{center}
\includegraphics[width=0.99\linewidth]{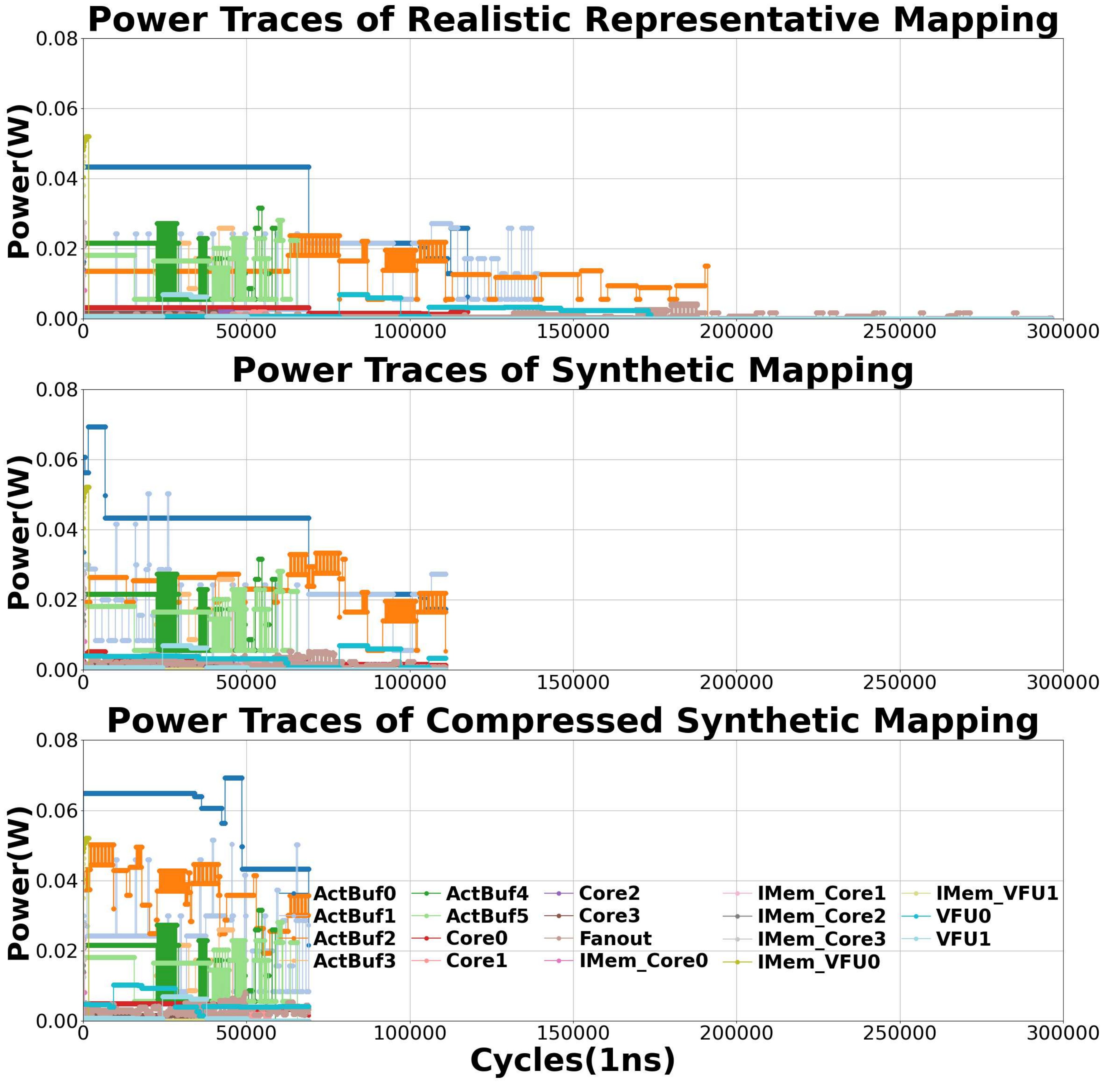}
\caption{Power traces of Resnet18 running with different mappings.}
\label{fig:compressedpower}
\end{center}
\end{figure}

First, we removed all the ``bubble" periods from the original {\em AERO's} power traces and obtained power traces representing the {\em AERO} mapping under ideal conditions. These traces are still representative of mapping and scheduling, as shown in the top subplot of Figure~\ref{fig:compressedpower}. Although these power traces ignore the back-and-forth relationship between different layers of the CNN model, they help us investigate the upper limit of temperature increase during the CNN inference. Next, we further compressed these representative power traces to two different degrees by superimposing two separate power traces (separated at 69,000 and 11,000 cycles) to obtain two synthetic power traces, as shown in the other subplots of Figure~\ref{fig:compressedpower}. This aggressive compression allows us to analyze the true upper limit of temperature increase during CNN inference, representing the upper bound of the transient temperature. This temperature can potentially be reached by real workloads in worst-case scenarios but can never be fully achieved.

\begin{figure}[!thbp]
\begin{center}
\includegraphics[width=0.99\linewidth]{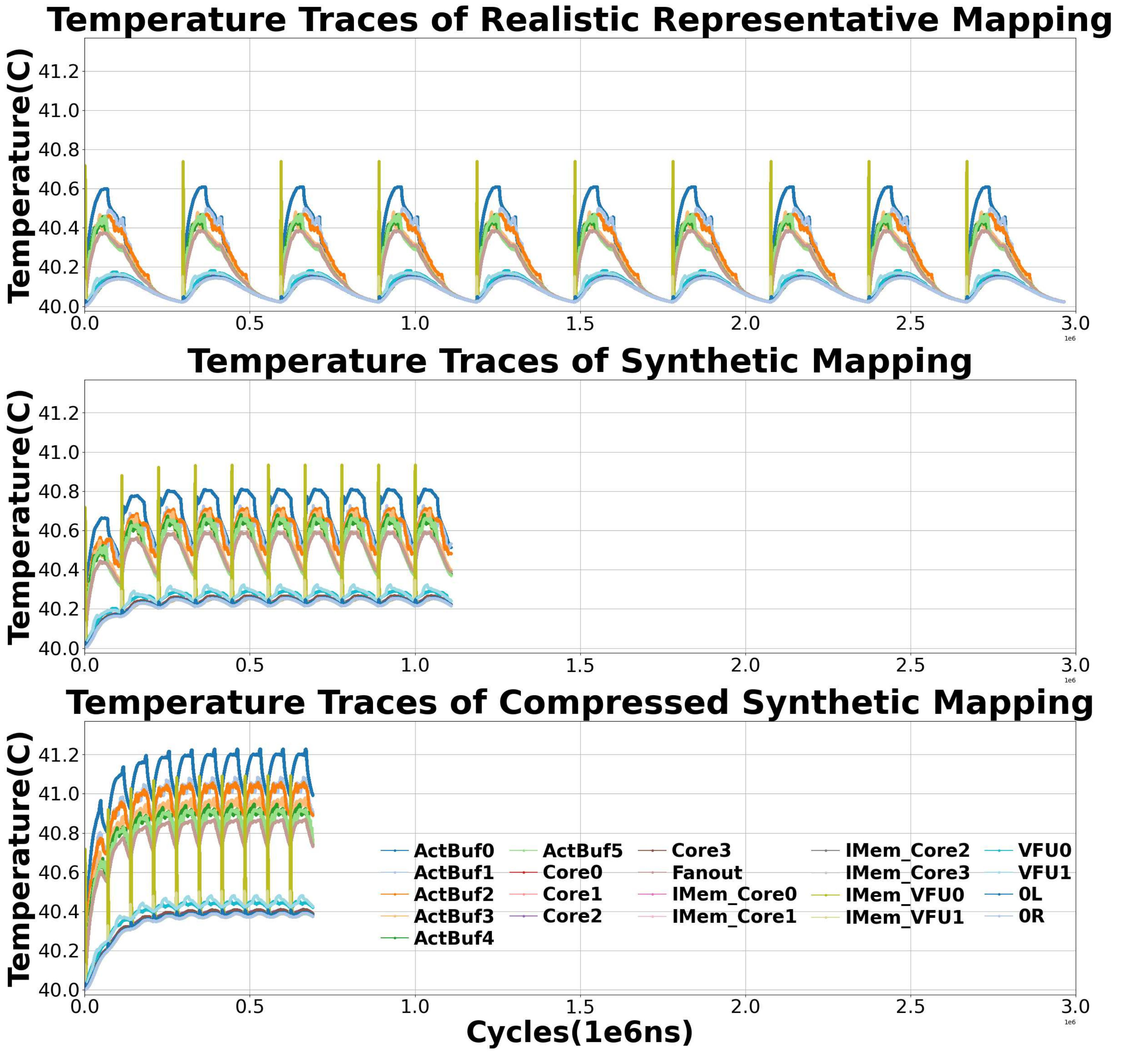}
\caption{Temperature traces of Resnet18 running with different mappings.}
\label{fig:compressedtemp}
\end{center}
\end{figure}

Figure~\ref{fig:compressedtemp} shows the transient temperature traces corresponding to the power traces of three different scenarios in Figure~\ref{fig:compressedpower}. These traces were run ten times repetitively, as shown in Figure~\ref{fig:compressedtemp}. The instantaneous temperature peak increases as we compress the power traces to focus on performance and reduce the inactive period. However, after running the workload two times, the temperature equilibrium between the internal and external environments is approached, and the further increase in temperature is not significant.

\section{Conclusion}
In this paper, we presented \safepath, a system-level methodology for efficient power and thermal estimation of CNN accelerators. \safepath integrates hardware mapping, power calculation, and temperature analysis to provide critical insights into the thermal behavior of architectures. Demonstrated on the TANIA architecture, \safepath accurately estimates power and temperature in both steady-state and transient scenarios. Future work will extend \safepath to multi-tile and multi-cluster scenarios, enhancing its applicability and comprehensiveness. Additionally, we will test the methodology on various CNN accelerators to validate its effectiveness across different architectures, ensuring its robustness and reliability in diverse contexts. \label{sec:conclusion}

\bibliographystyle{IEEEtran}
\bibliography{references.bib}

\begin{thebibliography}{10}
\providecommand{\url}[1]{#1}
\csname url@samestyle\endcsname
\providecommand{\newblock}{\relax}
\providecommand{\bibinfo}[2]{#2}
\providecommand{\BIBentrySTDinterwordspacing}{\spaceskip=0pt\relax}
\providecommand{\BIBentryALTinterwordstretchfactor}{4}
\providecommand{\BIBentryALTinterwordspacing}{\spaceskip=\fontdimen2\font plus
\BIBentryALTinterwordstretchfactor\fontdimen3\font minus \fontdimen4\font\relax}
\providecommand{\BIBforeignlanguage}[2]{{%
\expandafter\ifx\csname l@#1\endcsname\relax
\typeout{** WARNING: IEEEtran.bst: No hyphenation pattern has been}%
\typeout{** loaded for the language `#1'. Using the pattern for}%
\typeout{** the default language instead.}%
\else
\language=\csname l@#1\endcsname
\fi
#2}}
\providecommand{\BIBdecl}{\relax}
\BIBdecl

\bibitem{capra2020updated}
M.~Capra, B.~Bussolino, A.~Marchisio, and et~al., ``An updated survey of efficient hardware architectures for accelerating deep convolutional neural networks,'' \emph{Future Internet}, vol.~12, no.~7, p. 113, 2020.

\bibitem{chen2022thermal}
K.-C. Chen, H.-W. Tang, C.-H. Wu, and C.-H. Chen, ``Thermal sensor placement for multicore systems based on low-complex compressive sensing theory,'' \emph{IEEE TCAD}, vol.~41, no.~11, pp. 5100--5111, 2022.

\bibitem{chen2022wrap}
P.-Y. Chen, F.-Y. Gu, Y.-H. Huang, and C.~Lin, ``Wrap: Weight remapping and processing in rram-based neural network accelerators considering thermal effect,'' in \emph{2022 Design, Automation \& Test in Europe Conference \& Exhibition (DATE)}.\hskip 1em plus 0.5em minus 0.4em\relax IEEE, 2022, pp. 1245--1250.

\bibitem{sze2020evaluate}
V.~Sze, Y.-H. Chen, T.-J. Yang, and J.~S. Emer, ``How to evaluate deep neural network processors: Tops/w (alone) considered harmful,'' \emph{IEEE Solid-State Circuits Magazine}, vol.~12, no.~3, pp. 28--41, 2020.

\bibitem{song2017pipelayer}
L.~Song, X.~Qian, H.~Li, and Y.~Chen, ``Pipelayer: A pipelined reram-based accelerator for deep learning,'' in \emph{IEEE international symposium on high performance computer architecture (HPCA)}.\hskip 1em plus 0.5em minus 0.4em\relax IEEE, 2017, pp. 541--552.

\bibitem{shafiee2016isaac}
A.~Shafiee, A.~Nag, N.~Muralimanohar, and et~al., ``{ISAAC}: A convolutional neural network accelerator with in-situ analog arithmetic in crossbars,'' \emph{ACM SIGARCH Computer Architecture News}, vol.~44, no.~3, pp. 14--26, 2016.

\bibitem{yang2022aero}
S.~Yang, D.~Bhattacharjee, V.~B. Kumar, and et~al., ``{AERO}: Design space exploration framework for resource-constrained {CNN} mapping on tile-based accelerators,'' \emph{IEEE Journal on Emerging and Selected Topics in Circuits and Systems}, 2022.

\bibitem{cosemans2019towards}
S.~Cosemans, B.~Verhoef, J.~Doevenspeck, and et~al., ``Towards 10000tops/w dnn inference with analog in-memory computing--a circuit blueprint, device options and requirements,'' in \emph{2019 IEEE International Electron Devices Meeting (IEDM)}.\hskip 1em plus 0.5em minus 0.4em\relax IEEE, 2019, pp. 22--2.

\bibitem{zervakis2022thermal}
G.~Zervakis, I.~Anagnostopoulos, S.~Salamin, O.~Spantidi, I.~Roman-Ballesteros, J.~Henkel, and H.~Amrouch, ``Thermal-aware design for approximate dnn accelerators,'' \emph{IEEE Transactions on Computers}, vol.~71, no.~10, pp. 2687--2697, 2022.

\bibitem{shukla2021temperature}
P.~Shukla, S.~S. Nemtzow, V.~F. Pavlidis, E.~Salman, and A.~K. Coskun, ``Temperature-aware optimization of monolithic 3d deep neural network accelerators,'' in \emph{Proceedings of the 26th Asia and South Pacific Design Automation Conference}, 2021, pp. 709--714.

\bibitem{zhang2021full}
J.~Zhang, S.~Sadiqbatcha, M.~O’Dea, H.~Amrouch, and S.~X.-D. Tan, ``Full-chip power density and thermal map characterization for commercial microprocessors under heat sink cooling,'' \emph{IEEE TCAD}, vol.~41, no.~5, pp. 1453--1466, 2021.

\bibitem{huang2006hotspot}
W.~Huang, S.~Ghosh, S.~Velusamy, K.~Sankaranarayanan, K.~Skadron, and M.~R. Stan, ``Hotspot: A compact thermal modeling methodology for early-stage {VLSI} design,'' \emph{IEEE Transactions on very large scale integration (VLSI) systems}, vol.~14, no.~5, pp. 501--513, 2006.

\bibitem{chen2018systemc}
Y.~Chen, S.~Vinco, E.~Macii, and M.~Poncino, ``{SystemC-AMS} thermal modeling for the co-simulation of functional and extra-functional properties,'' \emph{ACM TODAES}, vol.~24, no.~1, pp. 1--26, 2018.

\bibitem{jiang2022chip}
L.~Jiang, A.~Dowling, Y.~Liu, and M.-C. Cheng, ``Chip-level thermal simulation for a multicore processor using a multi-block model enabled by proper orthogonal decomposition,'' in \emph{iTherm}.\hskip 1em plus 0.5em minus 0.4em\relax IEEE, 2022, pp. 1--7.

\bibitem{sridhar20133d}
A.~Sridhar, A.~Vincenzi, D.~Atienza, and T.~Brunschwiler, ``{3D-ICE}: A compact thermal model for early-stage design of liquid-cooled {ICs},'' \emph{IEEE Transactions on Computers}, vol.~63, no.~10, pp. 2576--2589, 2013.

\bibitem{siddhu2022comet}
L.~Siddhu, R.~Kedia, S.~Pandey, M.~Rapp, A.~Pathania, J.~Henkel, and P.~R. Panda, ``{CoMeT}: An integrated interval thermal simulation toolchain for {2D, 2.5 D, and 3D} processor-memory systems,'' \emph{ACM TACO}, vol.~19, no.~3, pp. 1--25, 2022.

\bibitem{wu2019accelergy}
Y.~N. Wu, J.~S. Emer, and V.~Sze, ``Accelergy: An architecture-level energy estimation methodology for accelerator designs,'' in \emph{2019 IEEE/ACM International Conference on Computer-Aided Design (ICCAD)}.\hskip 1em plus 0.5em minus 0.4em\relax IEEE, 2019, pp. 1--8.

\bibitem{jayapala2005clustered}
M.~Jayapala, F.~Barat, T.~Vander~Aa, and et~al., ``Clustered loop buffer organization for low energy vliw embedded processors,'' \emph{IEEE Transactions on Computers}, vol.~54, no.~6, pp. 672--683, 2005.

\end{thebibliography}

\end{document}